\pdfoutput=1
\documentclass[11pt]{article}

\usepackage{ACL2023}

\usepackage{times}
\usepackage{latexsym}

\usepackage[T1]{fontenc}
\usepackage[utf8]{inputenc}

\usepackage{microtype}

\usepackage{amsfonts,amssymb,amsmath,amsthm,amsopn,mathrsfs,mathtools,wasysym}	% math related
\usepackage{hhline,booktabs,colortbl,multirow,tabularx,diagbox,tablefootnote,threeparttable} % table related
\usepackage[ruled,linesnumbered,vlined,commentsnumbered]{algorithm2e}
\usepackage{graphicx}
\usepackage{subfig}
\usepackage{prettyref}
\usepackage{url}
\usepackage{tikz}
\usepackage{makecell}
\usepackage{color}

\usepackage{enumitem}

\definecolor{mycyan}{gray}{.7}

\def\our{\texttt{\textsc{BoostAug}}}
\newcommand{\pref}{\prettyref}

\DeclareMathOperator*{\argmax}{argmax}

\newrefformat{fig}{Figure~\ref{#1}}
\newrefformat{tab}{Table~\ref{#1}}
\newrefformat{sec}{Section~\ref{#1}}
\newrefformat{app}{Appendix~\ref{#1}}
\newrefformat{alg}{Algorithm~\ref{#1}}
\newrefformat{property}{Property~\ref{#1}}
\newrefformat{theorem}{Theorem~\ref{#1}}
\newrefformat{corollary}{Corollary~\ref{#1}}
\newrefformat{proposition}{Proposition~\ref{#1}}
\newrefformat{def}{Definition~\ref{#1}}
\newrefformat{eq}{equation~(\ref{#1})}

\title{Boosting Text Augmentation via Hybrid Instance Filtering Framework}

\author{
	Heng Yang, Ke Li\thanks{~~Corresponding author} \\ 
	Department of Computer Science, University of Exeter, EX4 4QF, Exeter, UK \\
	\texttt{\{hy345, k.li\}@exeter.ac.uk} \\
}

\begin{document}
\maketitle
\begin{abstract}
Text augmentation is an effective technique for addressing the problem of insufficient data in natural language processing. However, existing text augmentation methods tend to focus on few-shot scenarios and usually perform poorly on large public datasets. Our research indicates that existing augmentation methods often generate instances with shifted feature spaces, which leads to a drop in performance on the augmented data (for example, EDA generally loses $\approx 2\%$ in aspect-based sentiment classification). To address this problem, we propose a hybrid instance-filtering framework (\our) based on pre-trained language models that can maintain a similar feature space with natural datasets. \our\ is transferable to existing text augmentation methods (such as synonym substitution and back translation) and significantly improves the augmentation performance by $\approx 2-3\%$ in classification accuracy. Our experimental results on three classification tasks and nine public datasets show that \our\ addresses the performance drop problem and outperforms state-of-the-art text augmentation methods. Additionally, we release the code to help improve existing augmentation methods on large datasets.

\end{abstract}
	
\section{Introduction}
\label{sec:introduction}
\begin{figure}[t!]
\centering
\includegraphics[width=\linewidth]{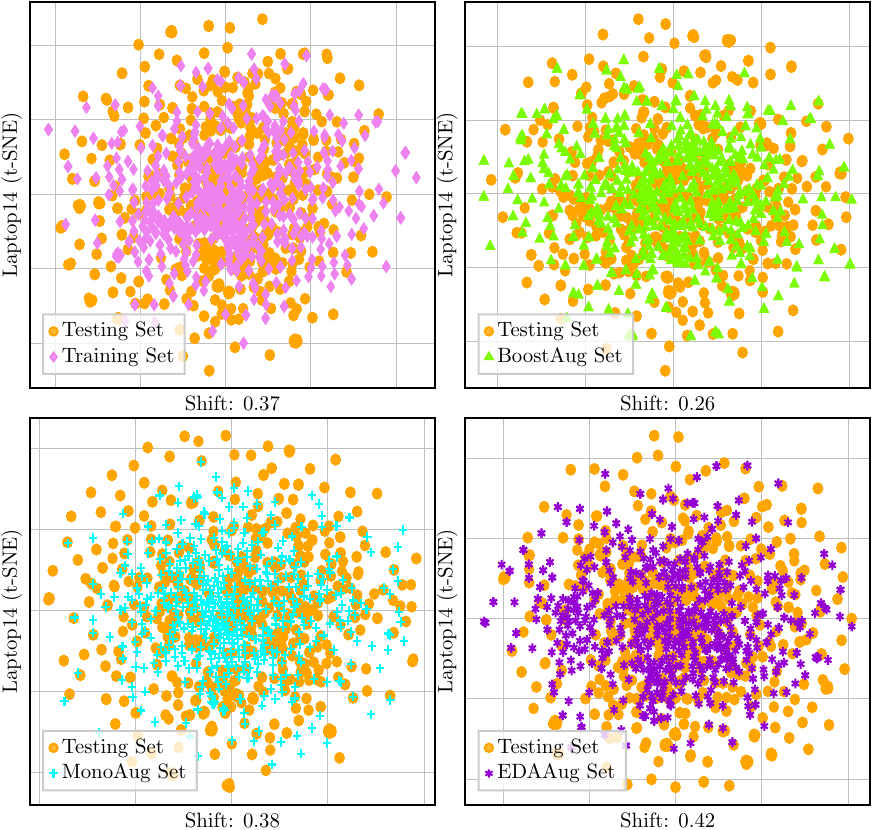}
\caption{The visualization of feature space shift of the Laptop14 dataset based on $t$-SNE. We calculate the shift metric $\mathcal{S}$ of feature space between augmented and natural instances. The augmentation methods are \texttt{BoostAug}, \texttt{MonoAug}, and \texttt{EDA} augmentation, respectively. Our \texttt{BoostAug} has the least feature space shift.}
\label{fig:rq3_sne}
\end{figure}

Recent pre-trained language models (PLMs) ~\cite{DevlinCLT19,BrownMRSKDNSSAA20,HeLGC21,YooPKLP21} have been able to learn from large amounts of text data. However, this also leads to a critical problem of data insufficiency in many low-resource fine-tuning scenarios \cite{ChenYY20,ZhouZTJY22,Miao0021,KimWOCH22,WangXSHTGJ22, YangDCSRD22}. Despite this, existing augmentation studies still encounter failures on large public datasets. While some studies\cite{NgCG20,BodyTLLZ21,ChangSZDS21,LuoLLZ21} have attempted to leverage the language modeling capabilities of PLMs in text augmentation, these methods still suffer from performance drops on large datasets.

To explore the root cause of this failure mode, we conducted experiments to explain the difference between ``good'' and ``bad'' augmentation instances. Our study found that existing augmentation methods ~\cite{WeiZ19,Coulombe18,LiJDLW19,KumarBBT19,NgCG20} usually fail to maintain the feature space in augmentation instances, which leads to bad instances. This shift in feature space occurs in both edit-based and PLM-based augmentation methods. For example, edit-based methods can introduce breaking changes that corrupt the meaning of the text, while PLM-based methods can introduce out-of-vocabulary words. In particular, for the edit-based methods, the shifted feature space mainly comes from breaking text transformations, such as changing important words (e.g., \lq \texttt{but} \rq\  ) in sentiment analysis. As for PLM-based methods, they usually introduce out-of-vocabulary words due to word substitution and insertion, which leads to an adverse meaning in sentiment analysis tasks.

To address the performance drop in existing augmentation methods caused by shifted feature space, we propose a hybrid instance-filtering framework (\our) based on PLMs to guide augmentation instance generation. Unlike other existing methods ~\citep{KumarCC20}, we use PLMs as a powerful instance filter to maintain the feature space, rather than as an augmentor. This is based on our finding that PLMs fine-tuned on natural datasets are familiar with the identical feature space distribution. The proposed framework consists of four instance filtering strategies: perplexity filtering, confidence ranking, predicted label constraint, and a cross-boosting strategy. These strategies are discussed in more detail in section ~\pref{sec:filtering}. Compared to prominent studies, \our\ is a pure instance-filtering framework that can improve the performance of existing text augmentation methods by maintaining the feature space.

With the mitigation of feature space shift, \our\ can generate more valid augmentation instances and improve existing augmentation methods' performance, which more augmentation instances generally trigger performance sacrifice in other studies ~\cite{Coulombe18,WeiZ19,LiJDLW19,KumarCC20}). According to our experimental results on three fine-grained and coarse-grained text classification tasks, \our\footnote{We release the source code and experiment scripts of \our\ at: \url{https://github.com/yangheng95/BoostTextAugmentation}.} significantly alleviates feature space shifts for existing augmentation methods.

Our main contributions are:
\begin{itemize}[leftmargin=*,noitemsep,nolistsep]
\item We propose the feature space shift to explain the performance drop in existing text augmentation methods, which is ubiquitous in full dataset augmentation scenarios.
\item We propose a universal augmentation instance filter framework to mitigate feature space shift and significantly improve the performance on the ABSC and TC tasks.
\item Our experiments show that the existing text augmentation methods can be easily improved by employing \our.
\end{itemize}

%!TeX root=main.tex
\section{Proposed Method}
\label{sec:method}

\begin{figure*}[t!]
\centering
\includegraphics[width=\textwidth]{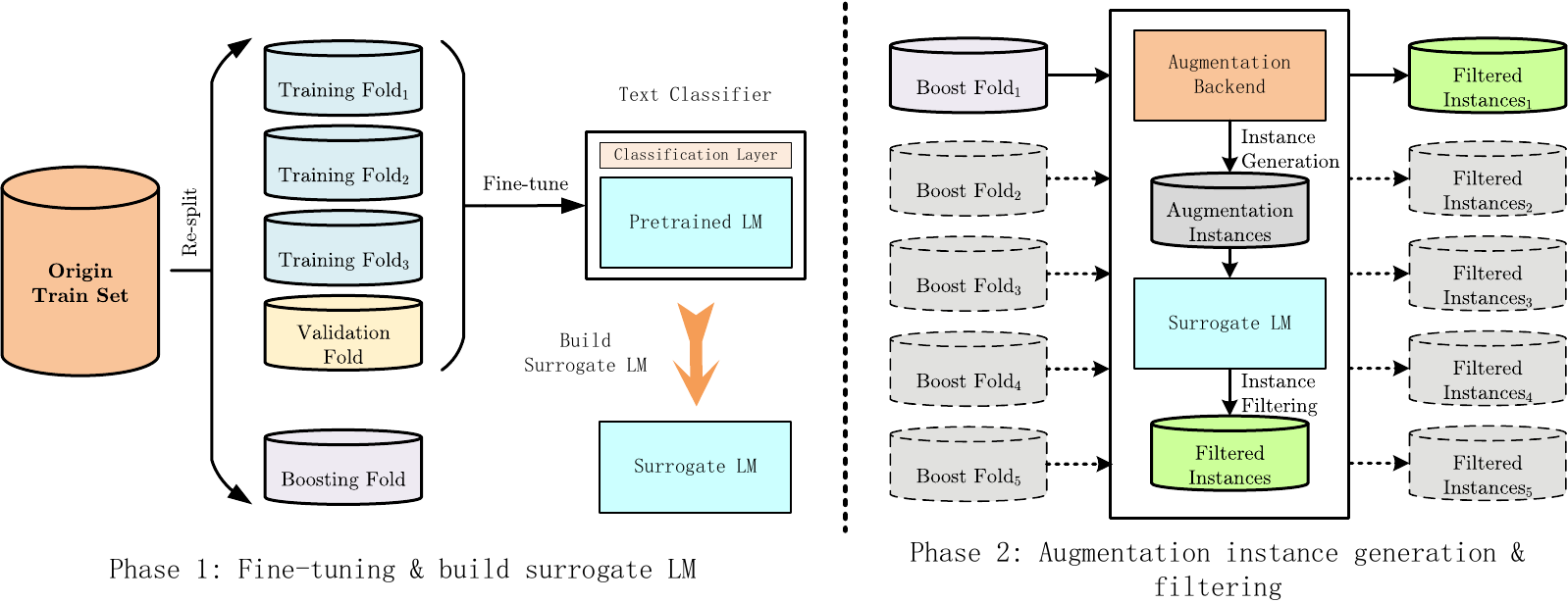}
\caption{The workflow of \our\ can be divided into two phases: \texttt{Phase \#$1$} and \texttt{Phase \#$2$}. In \texttt{Phase \#$1$}, we fine-tune a DeBERTa-based classification model using re-split training and validation sets and extract the fine-tuned DeBERTa to build a surrogate language model. In \texttt{Phase \#$2$}, \our\ employs a text augmentation backend to generate raw augmentations and filters out low-quality instances identified by the surrogate language model. To avoid data overlapping between the training folds and validation fold, \our\ performs $k$-fold cross-boosting, meaning that \texttt{Phase \#$1$} and \#$2$ are repeated $k$ times.}
\label{fig:workflow}
\end{figure*}

\begin{algorithm}[t!]
    \footnotesize
    Split $\mathcal{D}$ into $k$ folds, $\mathcal{D}:=\{\mathcal{F}^i\}_{i=1}^k$\;
    
    $\mathcal{D}_\mathrm{aug}:=\emptyset$\;
    \For{$i\leftarrow 1$ \KwTo $k$}{
        $\mathcal{D}_\mathrm{aug}^i:=\emptyset$, $\mathcal{D}_\mathrm{boost}^{i}:=\mathcal{F}^i$\;
        Randomly pick up $k-2$ folds except $\mathcal{F}^i$ to constitute $\mathcal{D}_\mathrm{train}^i$\;
        $\mathcal{D}_\mathrm{valid}^i:=\mathcal{F}\setminus(\mathcal{F}^i\bigcup\mathcal{D}_\mathrm{train}^i)$\;
        Use the \texttt{DeBERTa} on $\mathcal{D}_\mathrm{train}^i$ and $\mathcal{D}_\mathrm{valid}^i$ to build the surrogate language model\;
        \ForAll{$d_\mathrm{org}\in\mathcal{D}_\mathrm{boost}^{i}$}{
            $\mathcal{D}_\mathrm{aug}^i:=F(d_\mathrm{org}^i,\tilde{N},\Theta)$\;
            \ForAll{$d_\mathrm{aug}\in\mathcal{D}_\mathrm{aug}^i$}{
                Use the surrogate language model to predict $\mathbb{P}(d_\mathrm{aug})$, $\mathbb{C}(d_\mathrm{aug})$, and the $\tilde{\ell}_\mathrm{aug}$ of $d_\mathrm{aug}$\;
                \If{$\mathbb{P}(d_\mathrm{aug})\geq\alpha\ \|\ \mathbb{C}(d_\mathrm{aug})\leq\beta\ \|\ \tilde{\ell}_{d_{\mathrm{aug}}} \neq \tilde{\ell}_{d_{\mathrm{org}}}$}{
                    $\mathcal{D}_\mathrm{aug}^i:=\mathcal{D}_\mathrm{aug}^i\setminus\{d_\mathrm{aug}\}$\;
                }
            }
            $\mathcal{D}_\mathrm{aug}:=\mathcal{D}_\mathrm{aug}\bigcup\mathcal{D}_\mathrm{aug}^i$\;
        }
        $\mathcal{D}_\mathrm{aug}:=\mathcal{D}_\mathrm{aug}\bigcup\mathcal{D}_\mathrm{boost}^{i}$\;
    }
    \Return{$\mathcal{D}_\mathrm{aug}$} 
    \caption{The pseudo code of \our}
    \label{alg:boostaug}
\end{algorithm}
The workflow of \our\ is shown in \pref{fig:workflow} and the pseudo code is given in~\pref{alg:boostaug}. Different from most existing studies, which focus on unsupervised instance generation, \our\
serves as an instance filter to improve existing augmentation methods. The framework consists of two main phases: 1) \texttt{Phase \#$1$}: the training of surrogate language models; 2) \texttt{Phase \#$2$}: surrogate language models guided augmentation instance filtering. The following paragraphs will provide a detailed explanation of each step of the implementation.

\subsection{Surrogate Language Model Training}
\label{sec:surrogate_lm}

At the beginning of \texttt{Phase \#$1$}, the original training dataset is divided into $k\geq 3$ folds where the $k-2$ ones are used for training (denoted as the training fold) while the other two are used for the validation and augmentation purposes, denoted as the validation and boosting fold, respectively\footnote{We iteratively select the $i$-th fold, $i \in {1, \cdots, k}$, as the boosting fold (line $3$ in \pref{alg:boostaug}). The validation fold is used to select the best checkpoint of the surrogate language model to filter the augmented instances. This process is repeated $k$ times to ensure that all the folds have been used for validation and boosting at least once, thus avoiding data overlapping between the training and validation folds.} (lines $4$-$6$). Note that the generated augmentation instances, which will be introduced in \pref{sec:instance_generation}, can be identical to the instances in training folds the surrogate language model. This data overlapping problem will lead to a shifted feature space. We argue that the proposed $k$-fold augmentation approach, a.k.a. ``cross-boosting'', can alleviate the feature space shift of the augmentation instances, which will be validated and discussed in detail in \pref{sec:rq3}. The main crux of \texttt{Phase \#$1$} is to build a surrogate language model as a filter to guide the elimination of harmful and poor augmentation instances.

We construct a temporary classification model using the \texttt{DeBERTa}~\cite{HeLGC21} architecture. This model is then fine-tuned using the data in the $k-2$ training folds and the validation fold to capture the semantic features present in the data (line $7$). It is important to note that we do not use the original training dataset for this fine-tuning process. Once the fine-tuning is complete, the language model constructed from the \texttt{DeBERTa} classification model is then utilized as the surrogate language model in the instance filtering step in \texttt{Phase \#$2$} of \our.

This is different from existing works that use a pre-trained language model to directly generate augmentation instances. We clarify our motivation for this from the following two aspects.
\begin{itemize}[leftmargin=*,noitemsep,nolistsep]
\item In addition to modeling the semantic feature, the surrogate language model can provide more information that can be useful for the quality control of the augmentation instances, such as text perplexity, classification confidence, and predicted label.
\item Compared to the instance generation, we argue that the instance filtering approach can be readily integrated with any existing text augmentation approach.
\end{itemize}
\subsection{Augmentation Instance Generation}
\label{sec:instance_generation}

As a building block of \texttt{Phase \#$2$}, we apply some prevalent data augmentation approaches as the back end to generate the augmentation instances in \our\ (line $9$). More specifically, let $\mathcal{D}_\mathrm{org}:=\{d_\mathrm{org}^i\}_{i=1}^N$ be the original training dataset. $d_\mathrm{org}^i:=\langle s^i,\ell^i\rangle$ is a data instance where $s^i$ indicates a sentence and $\ell^i$ is the corresponding label, $i\in{1,\cdots,N}$. By applying the transformation function $F(\cdot,\cdot,\cdot)$ upon $d_\mathrm{org}^{i}$ as follows, we expect to obtain a set of augmentation instances $\mathcal{D}_\mathrm{aug}^i$ for $d_\mathrm{org}^{i}$:
\begin{equation}
\mathcal{D}_\mathrm{aug}^i:=F(d_\mathrm{org}^i,\tilde{N},\Theta),
\end{equation}
where $\tilde{N}\geq 1$ is used to control the maximum number of generated augmentation instances. In the end, the final augmentation set is constituted as $\mathcal{D}_\mathrm{aug}:=\bigcup_{i=1}^N\mathcal{D}_\mathrm{aug}^i$ (line $14$). Note that depending on the specific augmentation back end, there can be more than one strategy to constitute the transformation function. For example, \texttt{EDA} ~\cite{WeiZ19} has four transformation strategies, including synonym replacement, random insertion, random swap, and random deletion. $\Theta$ consists of the parameters associated with the transformation strategies of the augmentation back end, e.g., the percentage of words to be modified and the mutation probability of a word.

\subsection{Instance Filtering}
\label{sec:filtering}

Our preliminary experiments have shown that merely using data augmentation can be detrimental to the modeling performance, no matter how many augmentation instances are applied in the training process. In addition, our experiments in~\pref{sec:rq3} have shown a surprising feature space shift between the original data and the augmented instances in the feature space. To mitigate this issue, \our\ proposes an instance filtering approach to control the quality of the augmentation instances. It consists of three filtering strategies, including perplexity filtering, confidence ranking, and predicted label constraint, which will be delineated in the following paragraphs, respectively. Note that all these filtering strategies are built on the surrogate language model developed in \texttt{Phase} \#$1$ of \our\ (lines $12$ and $13$).
\subsubsection{Perplexity Filtering}
\label{sec:perplexity}

Text perplexity is a widely used metric to evaluate the modeling capability of a language model~\cite{ChenG99, Sennrich12}. Our preliminary experiments have shown that low-quality instances have a relatively high perplexity. This indicates that perplexity information can be used to evaluate the quality of an augmentation instance. Since the surrogate language model built in \texttt{Phase} \#$1$ is bidirectional, the text perplexity of an augmentation instance $d_\mathrm{aug}$ is calculated as:
\begin{equation}
\mathbb{P}(d_\mathrm{aug})=\prod_{i=1}^s p\left(w_i\mid w_1,\cdots,w_{i-1},w_{i+1},\cdots,w_s\right),
\end{equation}
where $w_{i}$ represents the token in the context. $s$ is the number of tokens in $d_\mathrm{aug}$ and $p\left(w_i\mid w_1,\cdots,w_{i-1},w_{i+1},\cdots,w_s\right)$ is the probability of $w_i$ conditioned on the preceding tokens, according to the surrogate language model, $i\in{1,\cdots,s}$. Note that $d_\mathrm{aug}$ is treated as a low-quality instance and is discarded if $\mathbb{P}(d_\mathrm{aug})\geq\alpha$ while $\alpha \geq 0$ is a predefined threshold.

\subsubsection{Confidence Ranking}
\label{sec:confidence_ranking}

We observe a significant feature space shift in the augmentation instances. These instances will be allocated with low confidences by the surrogate language model. In this case, we can leverage the classification confidence as a driver to control the quality of the augmentation instances.
However, it is natural that long texts can have way more augmentation instances than short texts, thus leading to the so-called unbalanced distribution. Besides, the confidence of most augmentation instances is $\geq 95\%$, which is not selective as the criterion for instance filtering.
To mitigate the unbalanced distribution in augmentation instances and make use of confidence, we develop a confidence ranking strategy to eliminate the redundant augmentation instances generated from long texts while retaining the rare instances having a relatively low confidence. More specifically, we apply a softmax operation on the output hidden state learned by the surrogate language model, denoted as $\mathbb{H}(d_\mathrm{aug})$, to evaluate the confidence of $d_\mathrm{aug}$ as:
\begin{equation}
\mathbb{C}(d_\mathrm{aug})=\argmax\bigg(\frac{\exp(\mathbb{H}_{d\mathrm{aug}})}{\sum_1^c\exp(\mathbb{H}_{d\mathrm{aug}})}\bigg),
\end{equation}
where $c$ is the number of classes in the original training dataset. To conduct the confidence ranking, $2\times\tilde{N}$ instances are generated at first, while only the top $\tilde{N}$ instances are selected to carry out the confidence ranking. By doing so, we expect to obtain a balanced augmentation dataset even when there is a large variance in the confidence predicted by the surrogate language model. After the confidence ranking, the augmentation instances with $\mathcal{C}_{d_\mathrm{aug}}\leq\beta$ are discarded while $\beta\geq 0$ is a fixed threshold.

\subsubsection{Predicted Label Constraint}
\label{sec:label_constraint}

Due to some breaking text transformation, text augmentation can lead to noisy data, e.g., changing a word "greatest" to "worst" in a sentence leads to an adverse label in a sentiment analysis task. Since the surrogate language model can predict the label of an augmentation instance based on its confidence distribution, we develop another filtering strategy that eliminates the augmentation instances whose predicted label $\tilde{\ell}_{d_{\mathrm{aug}}}$ is different from the ground truth. By doing so, we expect to mitigate the feature space bias.

\subsection{Feature Space Shift Metric}
\label{sec:metric}

To quantify the shift of the feature space, we propose an ensemble metric based on the overlapping ratio and distribution skewness of the $t$-SNE-based augmented instances' feature space.

The feature space overlapping ratio measures the diversity of the augmented instances. A larger overlapping ratio indicates that more natural instances have corresponding augmented instances. On the other hand, the distribution skewness measure describes the uniformity of the distribution of the augmented instances. A smaller distribution skewness indicates that the natural instances have approximately equal numbers of corresponding augmented instances.
To calculate the feature space shift, we first calculate the overlapping ratio and distribution skewness of the natural instances and their corresponding augmented instances. The feature space shift is calculated as follows:
\begin{equation}
\mathcal{S} = 1-\mathcal{O} + sk,
\end{equation}
where $\mathcal{O}$ and $sk$ are the feature space convex hull overlapping ratio and feature space distribution skewness, which will be introduced in the following subsections.

\subsubsection{Convex hull overlapping calculation}
To calculate the convex hull overlapping rate, we use the Graham Scan algorithm\footnote{\url{https://github.com/shapely/shapely}.}~\cite{Graham72} to find the convex hulls for the test set and target dataset in the $t$-SNE visualization, respectively.

Let $\mathcal{P}_{1}$ and $\mathcal{P}_{2}$ represent the convex hulls of two datasets in the $t$-SNE visualization; we calculate the overlapping rate as follows:
\begin{equation}
\mathcal{O} = \frac{\mathcal{P}_{1}\cap\mathcal{P}_{2}}{\mathcal{P}_{1}\cup\mathcal{P}_{2}},
\end{equation}
where $\cap$ and $\cup$ denote convex hull intersection and union operations, respectively. $\mathcal{O}$ is the overlapping rate between $\mathcal{P}_{1}$ and $\mathcal{P}_{2}$.

\subsubsection{Distribution skewness calculation}

The skewness of an example distribution is computed as follows:
\begin{equation}
sk=\frac{m_{3}}{m_{2}^{3 / 2}},
\end{equation}
\begin{equation}
m_{i}=\frac{1}{N} \sum_{n=1}^{N}(x_{n}-\bar{x})^{i},
\end{equation}
where $N$ is the number of instances in the distribution; $sk$ is the skewness of an example distribution. $m_{i}$ and $\bar{x}$ are the $i$-th central moment and mean of the example distribution, respectively. Because the $t$-SNE has two dimensions (namely $x$ and $y$ axes), we measure the global skewness of the target dataset (e.g., training set, augmentation set) by summarizing the absolute value of skewness on the $x$ and $y$ axes in $t$-SNE:
\begin{equation}
sk^{g}=|sk^{x}|+|sk^{y}|,
\end{equation}
where $sk^{g}$ is the global skewness of the target dataset; $sk^{x}$ and $sk^{y}$ are the skewness on the $x$ and $y$ axes, respectively.

By combining the convex hull overlapping ratio and distribution skewness, the proposed feature space shift metric offers a comprehensive view of how well the augmented instances align with the original data distribution. This metric can be used to evaluate the effectiveness of different data augmentation approaches, as well as to inform the fine-tuning process for better model performance.

\section{Experimental Setup}
\label{sec:setup}

\subsection{Datasets}
\label{sec:datasets}

Our experiments are conducted on three classification tasks: the sentence-level text classification (TC), the aspect-based sentiment classification (ABSC), and natural language inference (NLI). The datasets used for the TC task include SST2, SST5~\cite{SocherPWCMNP13} from the Stanford Sentiment Treebank, and AGNews10K\footnote{We use the first $10,000$ examples to build the AGNews10K dataset ($7,000$ for training, $1,000$ for validation, and $2,000$ for testing), which is large enough compared to other datasets.}\cite{ZhangZL15}. Meanwhile, the datasets used for the ABSC task are Laptop14, Restaurant14\cite{PontikiGPPAM14}, Restaurant15~\cite{PontikiGPMA15}, Restaurant16~\cite{PontikiGPAMAAZQ16}, and MAMS~\cite{JiangCXAY19}. The datasets\footnote{We select the first $1000$ training examples as the training set and keep the original validation/testing sets for experimental efficiency.} used for the NLI task are the SNLI~\cite{BowmanAPM15} and MNLI~\cite{WilliamsNB18} datasets, respectively. The split of these datasets is summarized in \pref{tab:datasets}. The commonly used \texttt{Accuracy} (i.e., \texttt{Acc}) and macro \texttt{F1} are used as the metrics for evaluating the performance of different algorithms following existing research~\cite{WangHZZ16,ZhouZTLY21}. Additionally, all experiments are repeated five times with different random seeds. Detailed information on the hyper-parameter settings and sensitivity tests of $\alpha$ and $\beta$ can be found in \pref{app:parameters}.

\begin{table}[t!]
    \centering
	\caption{The summary of experimental datasets for the text classification, aspect-based sentiment analysis and natural language inference tasks.}
    \resizebox{\linewidth}{!}{ 
    \begin{tabular}{c|c|c|c}
        \hline
        \textbf{Dataset} & \textbf{Training Set} & \textbf{Validation Set} & \textbf{Testing Set} \\
        \hline
        Laptop14 & $2328$  & $0$     & $638$\\
        \hline
        Restaurant14 & $3608$  & $0$     & $1120$\\
        \hline
        Restaurant15 & $1120$  & $0$     & $540$\\
        \hline
        Restaurant16 & $1746$  & $0$     & $615$\\
        \hline
        MAMS  & $11186$ & $1332$  & $1336$\\
        \hline
        SST2  & $6920$  & $872$   & $1821$\\
        \hline
        SST5  & $8544$  & $1101$  & $2210$\\
        \hline
        AGNews10K & $7000$ & $1000$ & $2000$ \\
        \hline
        SNLI  & $1000$  & $10000$  & $10000$\\
        \hline
        MNLI & $1000$ & $20000$ & $0$ \\
        \hline
    \end{tabular}
    }
    \label{tab:datasets}
\end{table}
\subsection{Augment Backends}
\label{sec:backends}

We use \our\ to improve five state-of-the-art baseline text augmentation methods, all of which are used as the text augmentation backend of \our. Please find the introductions of these baselines in \pref{app:backends_app} and refer to \pref{tab:rq4_results} for detailed performance of \our\ based on different backends.

We also compare \our\ enhanced \texttt{EDA} with the following text augmentation methods:
\begin{itemize}[leftmargin=*,noitemsep,nolistsep]
\item \texttt{EDA} (TextAttack\footnote{\url{https://github.com/QData/TextAttack}}) \cite{WeiZ19} performs text augmentation via random word insertions, substitutions, and deletions.
\item \texttt{SynonymAug} (NLPAug\footnote{\url{https://github.com/makcedward/nlpaug}}) \cite{NiuB18} replaces words in the original text with their synonyms. This method has been shown to be effective in improving the robustness of models on certain tasks.
\item \texttt{TAA} \cite{RenZLSZ21} is a Bayesian optimization-based text augmentation method. It searches augmentation policies and automatically finds the best augmentation instances.
\item \texttt{AEDA} \cite{KarimiR021} is based on the \texttt{EDA}, which attempts to maintain the order of the words while changing their positions in the context. Besides, it alleviates breaking changes such as critical deletions and improves the robustness.
\item \texttt{AMDA} \cite{SiZQLWLS21} linearly interpolates the representations of pairs of training instances, which has a diversified augmentation set compared to discrete text adversarial augmentation.
\end{itemize}

In our experiments, \texttt{LSTM}, \texttt{BERT-BASE}\cite{DevlinCLT19}, and \texttt{DeBERTa-BASE}\cite{HeLGC21} are used as the objective models for the TC task. \texttt{FastLCF} is an objective model available for the ABSC task.

%!TeX root=main.tex

\begin{table*}[t!]
	\centering
	\caption{The performance comparison between \our-enhanced \texttt{EDA} and baseline augmentation methods. The best and second-best metric values are highlighted in \textbf{bold} and \underline{underlined} faces, respectively. \texttt{None} indicates the vanilla version without using text augmentation. $^\dagger$ indicates that \our\ is significantly better than the backend according to the Wilcoxon's rank sum test at a $0.05$ significance level. ``-'' indicates that FastLCF is not available for text classification or the results are not considered due to resource limitation. }
	\label{tab:full_experiments}
	\resizebox{\linewidth}{!}{
		\begin{tabular}{c|c|c|c|c|c|c|c|c|c|c|c||c|c|c|c|c|c}
			\hline
			\multirow{1}[4]{*}{\textbf{Augmentation}} & \multirow{1}[4]{*}{\textbf{Model}} & \multicolumn{2}{c|}{\textbf{Laptop14}} & \multicolumn{2}{c|}{\textbf{Restaurant14}} & \multicolumn{2}{c|}{\textbf{Restaurant15}} & \multicolumn{2}{c|}{\textbf{Restaurant16}} & \multicolumn{2}{c||}{\textbf{MAMS}} & \multicolumn{2}{c|}{\textbf{SST2}} & \multicolumn{2}{c|}{\textbf{SST5}} & \multicolumn{2}{c}{\textbf{AGNews10K}} \\
			\cline{3-18}   &  & \texttt{Acc}   & \texttt{F1}    & \texttt{Acc}   & \texttt{F1}    & \texttt{Acc}   & \texttt{F1}    & \texttt{Acc}   & \texttt{F1}    & \texttt{Acc}   & \texttt{F1} & \texttt{Acc}   & \texttt{F1}    & \texttt{Acc}   & \texttt{F1}   & \texttt{Acc}   & \texttt{F1}\\
			%		\hline
			\hline
			\multirow{4}[2]{*}{\texttt{None}}
			& LSTM  & 71.32 & 65.45 & 77.54 & 66.89 & 78.61 & 58.54 & 87.40 & 64.41 & 56.96 & 56.18     & 84.36 & 84.36 & 45.29 & 44.61 & 87.60 & 87.36 \\
			& BERT  & 79.47 & 75.70  & 85.18 & 78.31 & 83.61 & 69.73 & 91.3  & 77.16 & 82.78 & 82.04    & 90.88 & 90.88 & 53.53 & 52.06 & 92.47 & 92.26 \\
			& DeBERTa  & 83.31 & 80.02 & 87.72 & 81.73 & 86.58 & 74.22 & 93.01 & 81.93 & 83.31 & 82.87  & \underline{95.07} & \underline{95.07} & \underline{56.47} & \underline{55.58} & 92.30 & 92.13 \\
			& FastLCF  & 83.23 & 79.68 & 88.5  & 82.7  & 87.74 & 73.69 & 93.69 & 81.66 & 83.46 & 82.88  & - & - & - & - & - & - \\
			%		\hline
			\hline
			\multirow{4}[2]{*}{\texttt{EDA}}
			& LSTM  & 68.65 & 62.09 & 76.18 & 62.41 & 76.30 & 56.88 & 85.59 & 61.78 & 56.59 & 55.33     & 84.79 & 84.79 & 43.85 & 43.85 & 87.72 & 87.46 \\
			& BERT  & 78.37 & 74.23 & 83.75 & 75.38 & 81.85 & 65.63 & 91.38 & 77.27 & 81.81 & 81.10     & 91.16 & 91.16 & 51.58 & 50.49 & 92.50 & 92.28 \\
			& DeBERTa  & 80.96 & 78.65 & 86.79 & 79.82 & 84.44 & 70.40 & 93.01 & 77.59 & 81.96 & 81.96  & 94.07 & 94.07 & 56.43 & 53.88 & 92.55 & 92.33 \\
			& FastLCF  & 81.97 & 79.57 & 87.68 & 81.52 & 86.39 & 72.51 & 93.17 & 78.96 & 82.19 & 81.63  & - & - & - & - & - & - \\
			%		\hline
			\hline
			\multirow{4}[2]{*}{\texttt{SpellingAug}}
			& LSTM  & 67.24 & 60.30 & 75.36 & 63.01 & 73.52 & 49.04 & 84.72 & 53.92 & 55.99 & 55.16    & 83.14 & 83.14 & 41.45 & 40.40 & 87.25 & 86.96 \\
			& BERT  & 73.59 & 69.11 & 82.54 & 73.18 & 79.63 & 62.32 & 89.76 & 74.74 & 81.89 & 81.42    & 91.00 & 91.00 & 52.26 & 50.90 & 92.42 & 92.22 \\
			& DeBERTa & 80.17 & 76.01 & 85.13 & 76.67 & 85.83 & 71.54 & 92.76 & 78.33 & 81.89 & 81.24  & 93.68 & 93.68 & 55.95 & 53.78 & 92.68 & 92.50\\
			& FastLCF & 79.62 & 74.81 & 86.03 & 78.73 & 87.41 & 75.14 & 92.60 & 75.27 & 82.19 & 81.66  & - & - & - & - & - & -\\
			%		\hline
			\hline
			\multirow{4}[2]{*}{\texttt{SplitAug}}
			& LSTM  & 62.98 & 56.53 & 73.43 & 58.57 & 70.19 & 45.71 & 83.93 & 54.41 & 56.74 & 55.34     & 84.29 & 84.29 & 44.00 & 42.10 & 87.23 & 87.01 \\
			& BERT  & 75.47 & 70.56 & 82.86 & 74.48 & 82.87 & 65.19 & 90.98 & 77.51 & 81.74 & 81.35     & 90.88 & 90.88 & 51.99 & 50.95 & 92.45 & 92.16 \\
			& DeBERTa  & 79.15 & 75.72 & 86.03 & 79.28 & 85.46 & 70.43 & 92.76 & 79.79 & 81.59 & 81.09  & 94.29 & 94.29 & 55.51 & 49.77  & 92.52 & 92.29 \\
			& FastLCF  & 81.82 & 78.46 & 86.34 & 78.36 & 86.67 & 70.87 & 93.09 & 76.50 & 82.07 & 81.53  & - & - & - & - & - & - \\
			%		\hline
			\hline
			\multirow{4}[2]{*}{\texttt{ContextualWordEmbsAug}\tnote{b}}
			& LSTM  & 67.40 & 61.57 & 75.62 & 62.13 & 74.44 & 51.67 & 84.98 & 58.67 & 56.06 & 55.10  & 83.14 & 83.14 & 44.07 & 42.03 & 87.53 & 87.24 \\
			& BERT  & 75.63 & 70.79 & 83.26 & 75.11 & 78.61 & 61.48 & 90.24 & 72.37 & 81.29 & 80.50     & 91.02 & 91.02 & 51.27 & 50.27 & 92.10 & 91.86 \\
			& DeBERTa  & 76.88 & 71.98 & 85.49 & 77.22 & 84.63 & 70.50 & 92.28 & 77.42 & 81.66 & 81.32  & 94.12 & 94.12 & 55.48 & 53.60 & \underline{92.80} & \underline{92.62}\\
			& FastLCF  & 79.08 & 74.61 & 85.62 & 76.88 & 84.91 & 70.06 & 91.38 & 76.27 & 81.89 & 81.09  & - & - & - & - & - & - \\
			\hline
			\multirow{4}[2]{*}{\texttt{BackTranslationAug}\tnote{b}}
			& LSTM     & 68.50 & 62.22 & 78.12 & 66.70 & 78.85 & 59.08 & 86.97 & 63.47 & - & - & - & - & - & - & - & - \\
			& BERT     & 79.94 & 76.19 & 85.54 & 78.51 & 84.42 & 72.05 & 92.02 & 85.78 & - & - & - & - & - & - & - & - \\
			& DeBERTa  & 84.17 & 81.15 & 88.93 & 83.54 & 89.42 & 78.67 & 93.97 & 80.52 & - & - & - & - & - & - & - & - \\
			& FastLCF   & 82.76 & 79.82 & \underline{89.46} & \underline{84.94} & 88.13 & 75.70 & \underline{94.14} & 81.82 & - & - & - & - & - & - & - & - \\
			%			             \hline
			%			             \multirow{1}[1]{}{\texttt{TAA}}
			%			             & BERT & - & - & - & - & - & - & - & - & - & - & 90.94 & 90.94 & 52.55 & 51.29  & & \\
			\hline
			\hline
			\multirow{4}[2]{*}{\our~(\texttt{EDA})}
			& LSTM     & 73.20$^\dagger$ & 67.46$^\dagger$ & 79.12$^\dagger$ & 68.07$^\dagger$ & 80.06$^\dagger$ & 59.61$^\dagger$ & 87.80$^\dagger$ & 65.33$^\dagger$ & 59.21$^\dagger$ & 59.58$^\dagger$      & 85.83$^\dagger$ & 85.83$^\dagger$ & 45.93$^\dagger$ & 43.59$^\dagger$ & 88.45 & 88.16\\
			& BERT     & 80.10$^\dagger$ & 76.48$^\dagger$ & 86.34$^\dagger$ & 79.99$^\dagger$ & 86.12$^\dagger$ & 73.79$^\dagger$ & 91.95$^\dagger$ & 79.12$^\dagger$ & 84.01$^\dagger$ & \underline{83.44}$^\dagger$ & 92.33$^\dagger$ & 92.33$^\dagger$ & 53.94$^\dagger$ & 52.80$^\dagger$ & 92.48 & 92.25 \\
			& DeBERTa  & \underline{84.56}$^\dagger$ & \underline{81.77}$^\dagger$ & 89.02$^\dagger$ & 83.35$^\dagger$ & \underline{88.33}$^\dagger$ & \underline{76.77}$^\dagger$ & 93.58$^\dagger$ & \underline{81.93}$^\dagger$ & \textbf{84.51}$^\dagger$ & \textbf{83.97}$^\dagger$    & \textbf{96.09}$^\dagger$	& \textbf{96.09}$^\dagger$ & \textbf{57.78}$^\dagger$ & \textbf{56.15}$^\dagger$ & \textbf{92.95} & \textbf{92.76} \\
			& FastLCF  & \textbf{85.11}$^\dagger$ & \textbf{82.18}$^\dagger$ & \textbf{90.38}$^\dagger$ & \textbf{85.04}$^\dagger$ & \textbf{89.81}$^\dagger$ & \textbf{77.92}$^\dagger$ & \textbf{94.37}$^\dagger$ & \textbf{82.67}$^\dagger$ & \underline{84.13}$^\dagger$ & 82.97$^\dagger$    & - & - & - & - & - & -\\
			\hline
		\end{tabular}
	}
\end{table*}
\section{Experimental Results}
\label{sec:results}

\subsection{Main Results}
\label{sec:rq1}
From the results shown in Table~\ref{tab:full_experiments}, it is clear that \our\ consistently improves the performance of the text augmentation method \texttt{EDA} across all datasets and models. It is also worth noting that some traditional text augmentation methods can actually harm the performance of the classification models. Additionally, the performance improvement is relatively small for larger datasets like \texttt{SST-2}, \texttt{SST-5}, and \texttt{MAMS}. Furthermore, the performance of \texttt{LSTM} is more affected by text augmentation, as it lacks the knowledge gained from the large-scale corpus that is available in PLMs.

When comparing the different text augmentation methods, it is apparent that \texttt{EDA} performs the best, despite being the simplest method. On the other hand, \texttt{SplitAug} performs the worst for \texttt{LSTM} because its augmentation instances are heavily biased in the feature space due to the word splitting transformation. The performance of \texttt{SpellingAug} is similar to \texttt{EDA}. This can be attributed to the fact that PLMs have already captured some common misspellings during pre-training. Additionally, PLM-based augmentation methods like \texttt{WordsEmbsAug} tend to generate instances with unknown words, further exacerbating the feature space shift of the augmented texts.

\begin{table}[htbp]
	\centering
	\caption{The performance comparison on augmented \textbf{SST2} dataset between different augmentation methods. We list the standard deviations for each method, while ``-'' indicates the standard deviation is not available. $^*$ is derived from our experiments.}
	\resizebox{.9\linewidth}{!}{
	\begin{tabular}{c|c|c|c}
		\hline
		\textbf{Augmentation} & \textbf{Model} & \texttt{Acc}  & \texttt{F1}  \\
		\hline
		\texttt{None}$^*$  & BERT  & 90.88 (0.31) & 90.87 (0.31)\\
		\hline
		\texttt{EDA}$^*$ & BERT  & 90.99 (0.46) & 90.99 (0.46)\\
		\hline
  	\texttt{SynonymAug}$^*$  & BERT  & 91.32 (0.55)  & 91.31 (0.55)\\
		\hline
		\texttt{TAA}$^*$  & BERT  & 90.94 (0.31)  & 90.94 (0.31)\\
		\hline
		\texttt{AEDA}  & BERT  & 91.76 (~-~~)  & --- \\
		\hline
  	\texttt{AMDA}  & BERT  & 91.54 (~-~~)  & ---  \\
		\hline
		\our (\texttt{EDA}) & BERT  & \textbf{92.33} (0.29) & \textbf{92.33} (0.29) \\
		\hline
	\end{tabular}%
	}
	\label{tab:sota}%
\end{table}%
We also compare the performance of \our\ with several state-of-the-art text augmentation methods. The results of these comparisons can be found in Table~\ref{tab:sota}. From the results, it can be seen that even when using \texttt{EDA} ~\cite{WeiZ19} as the backend, \our\ outperforms other state-of-the-art methods such as \texttt{AEDA} ~\cite{KarimiR021}, \texttt{AMDA} ~\cite{SiZQLWLS21}, and Bayesian optimization-based \texttt{TAA} ~\cite{RenZLSZ21} on the full \texttt{SST2} dataset.

\subsection{Ablation Study}
\label{sec:rq2}

\begin{table*}[t!]
    \centering
    \caption{The performance comparison between different ablated variants of \our.}
    \label{tab:rq2_results}
    \resizebox{.8\linewidth}{!}{
       \begin{tabular}{c|c|c|c|c|c|c|c|c|c}
            \hline
            \multirow{1}[4]{*}{\textbf{Ablation}} & \multirow{1}[4]{*}{\textbf{Model}} & \multicolumn{2}{c|}{\textbf{MAMS}} & \multicolumn{2}{c|}{\textbf{SST2}} & \multicolumn{2}{c}{\textbf{SST5}}  & \multicolumn{2}{c}{\textbf{AGNews10K}} \\
            \cline{3-10}          &       & \texttt{Acc}   & \texttt{F1}    & \texttt{Acc}   & \texttt{F1}    & \texttt{Acc}   & \texttt{F1} & \texttt{Acc}   & \texttt{F1} \\
            \hline
            \multirow{2}[3]{*}{\our (\texttt{EDA})}
            & \texttt{LSTM} & 59.21 & 59.58 & 85.77 & 85.77 &  45.79 & 43.84 & 88.45 & 88.16 \\
            & \texttt{BERT} & 84.01 & 83.44 & 92.33 & 92.33 & 52.38 & 51.70  & 92.48 & 92.25 \\
            & \texttt{DeBERTa}   & \textbf{84.51} & \textbf{83.97} & \textbf{96.09}& \textbf{96.09} & 57.78 & 56.15 & \textbf{92.95} & \textbf{92.76} \\
            \hline
            \hline
            \multirow{2}[3]{*}{\texttt{MonoAug}}
            & \texttt{LSTM}  & 57.26 & 55.70 & 84.62 & 84.62 & 45.25 & 42.91 & 87.55 & 87.32 \\
            & \texttt{BERT}  & 83.68 & 83.04 & 91.16 & 91.16 & 52.90 & 52.12 & 92.40 & 92.19 \\
            & \texttt{DeBERTa}  & 83.38 & 82.87 & 94.18 & 94.18 & 56.92 & 55.81 & 92.32 & 92.09 \\
            \hline
            \multirow{2}[3]{*}{\texttt{w/o Confidence}}
            & \texttt{LSTM}  & 56.19 & 55.94 & 85.08 & 85.08 & 45.48 & 44.89 & 86.98 & 86.61 \\
            & \texttt{BERT}  & 83.26 & 82.62 & 91.16 & 91.16 & 53.21 & 52.27 & 92.20 & 92.00 \\
            & \texttt{DeBERTa}  & 83.47 & 82.08 & 95.22 & 95.22 & 57.10 & 55.97 & 92.93 & 92.75 \\
            \hline
            \multirow{2}[3]{*}{\texttt{w/o Perplexity}}
            & \texttt{LSTM}  & 55.54 & 55.46 & 85.67 & 85.67 & 46.47 & 43.25 & 87.40 & 86.99 \\
            & \texttt{BERT}  & 83.16 & 82.58 & 92.04 & 92.04 & 52.67 & 51.02 & 92.50 & 92.30 \\
            & \texttt{DeBERTa} & 83.53 & 83.04 & 95.39 & 95.39 & \textbf{58.10} & \textbf{56.78} & 92.60 & 92.36\\
            \hline
            \multirow{2}[3]{*}{\texttt{w/o Label Constraint}}
            & \texttt{LSTM}  & 56.06 & 55.00 & 84.90 & 84.88 & 44.75 & 43.44 & 86.55 & 86.23 \\
            & \texttt{BERT}  & 83.01 & 82.57 & 92.04 & 92.03 & 52.58 & 51.33 & 91.80 & 91.60\\
            & \texttt{DeBERTa} & 82.41 & 82.01 & 95.06 & 95.06 & 56.70 & 54.91 & 92.85 & 92.60 \\
            \hline
        \end{tabular}%
    }
\end{table*}

To gain a deeper understanding of the working mechanism of \our, we conduct experiments to evaluate the effectiveness of cross-boosting, predicted label constraint, confidence ranking, and perplexity filtering. The results, which can be found in~\pref{tab:rq2_results}, show that the performance of the variant \texttt{MonoAug} is significantly lower than that of \our. This is because \texttt{MonoAug} trains the surrogate language model using the entire training set, leading to a high degree of similarity between the original and augmentation instances. This data overlapping problem, as discussed in~\pref{sec:surrogate_lm}, results in biased instance filtering and overfitting of the instances to the training fold data distribution. Additionally, the variant without the perplexity filtering strategy performs the worst, indicating that the perplexity filtering strategy is crucial in removing instances with syntactical and grammatical errors. The performance of the variants without the predicted label constraint and confidence ranking is similar, with the label constraint helping to prevent the mutation of features into an adverse meaning and the confidence ranking helping to eliminate out-of-domain words and reduce feature space shift.

\subsection{Feature Space Shift Investigation}
\label{sec:rq3}

In this subsection, we explore the feature space shift problem in more detail by using visualizations and the feature space shift metric. We use $t$-SNE to visualize the distribution of the features of the testing set and compare it to different augmented variants. The full results of feature space shift metrics are available in \pref{fig:app_tsne}. The results of feature space shift metrics in our experiment show that the augmentation instances generated by \our\ have the least shift of feature space. Specifically, the overlapping ratio and skewness in relation to the testing set are consistently better than those of the training set. This explains the performance improvement seen when using \our\ in previous experiments. In contrast, the augmentation instances generated by \texttt{EDA}, which was the best peer text augmentation method, have a worse overlapping rate compared to even the training set. This explains the performance degradation when using \texttt{EDA} on the baseline classification models. It is also noteworthy that the quality of the augmentation instances generated by \texttt{MonoAug} is better than \texttt{EDA}.

\subsection{Effect of Augmentation Instances Number}
\label{sec:rq5}

\begin{figure*}[t!]
    \centering
    \includegraphics[width=.9\linewidth]{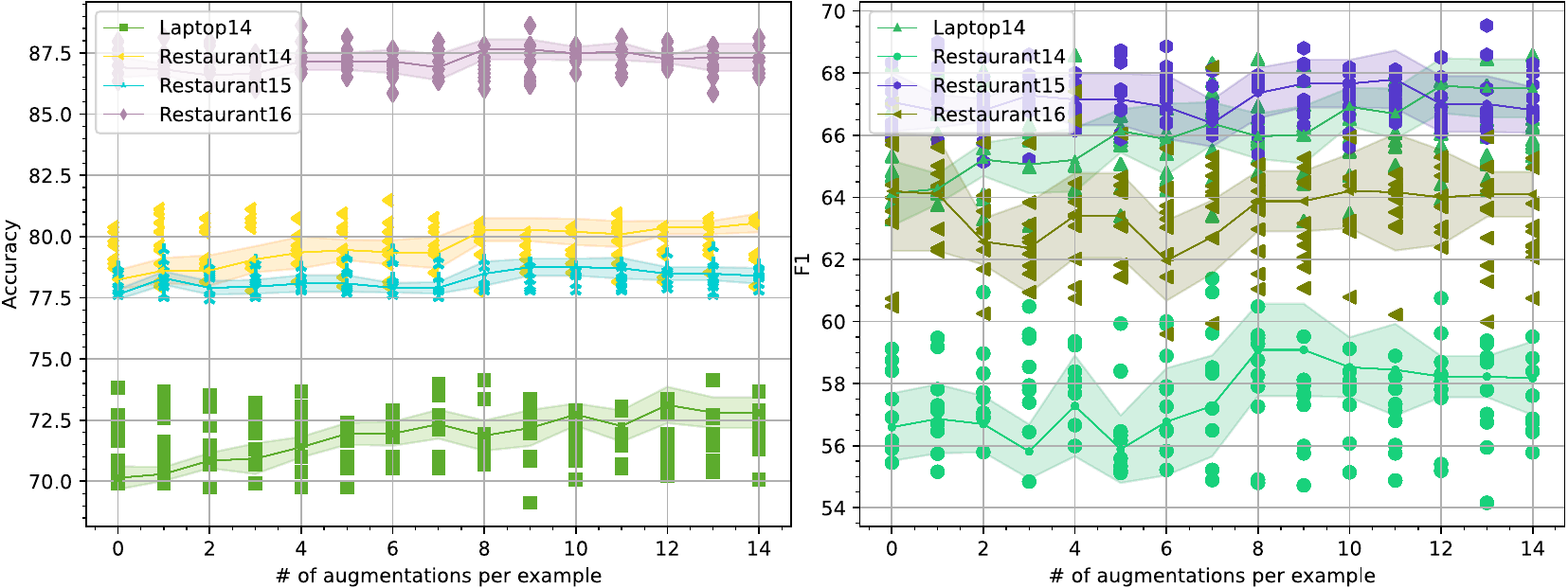}
    \caption{Trajectories of the \texttt{Acc} and the \texttt{F1} values with error bars versus the number of augmentation instances generated for an example by using \our (\texttt{EDA}). The trajectory visualization plot of \texttt{MonoAug} and \texttt{EDA} can be found in \pref{fig:rq4_full}}
    \label{fig:rq4_boostaug}
\end{figure*}
To further understand the effectiveness of \our, we conduct an experiment to analyze the relationship between the number of augmentation instances generated and the performance of the classification models. We use \texttt{Acc} and \texttt{F1} as the evaluation metrics and plot the trajectories of these metrics with error bars against the number of augmentation instances generated for an example by using \our. The results are shown in \pref{fig:rq4_boostaug}. For comparison, the trajectory visualization plots of \texttt{MonoAug} and \texttt{EDA} can also be found in \pref{fig:rq4_full}. From the results, it is clear to see that the performance of the classification models improves as the number of augmentation instances increases, but eventually reaches a saturation point. Furthermore, it is observed that the performance improvement achieved by \our\ is consistently better than that of \texttt{MonoAug} and \texttt{EDA}. This further confirms the effectiveness of \our\ in mitigating the feature space shift problem and improving the performance of the classification models.

However, it is also important to consider the computational budgets required to generate a large number of augmentation instances, as this can impact the overall efficiency of the text augmentation method being used.

\subsection{Hyper-parameter Sensitivity Analysis}
\label{sec:param}
\begin{figure}[htbp]
    \centering
    \includegraphics[width=\linewidth]{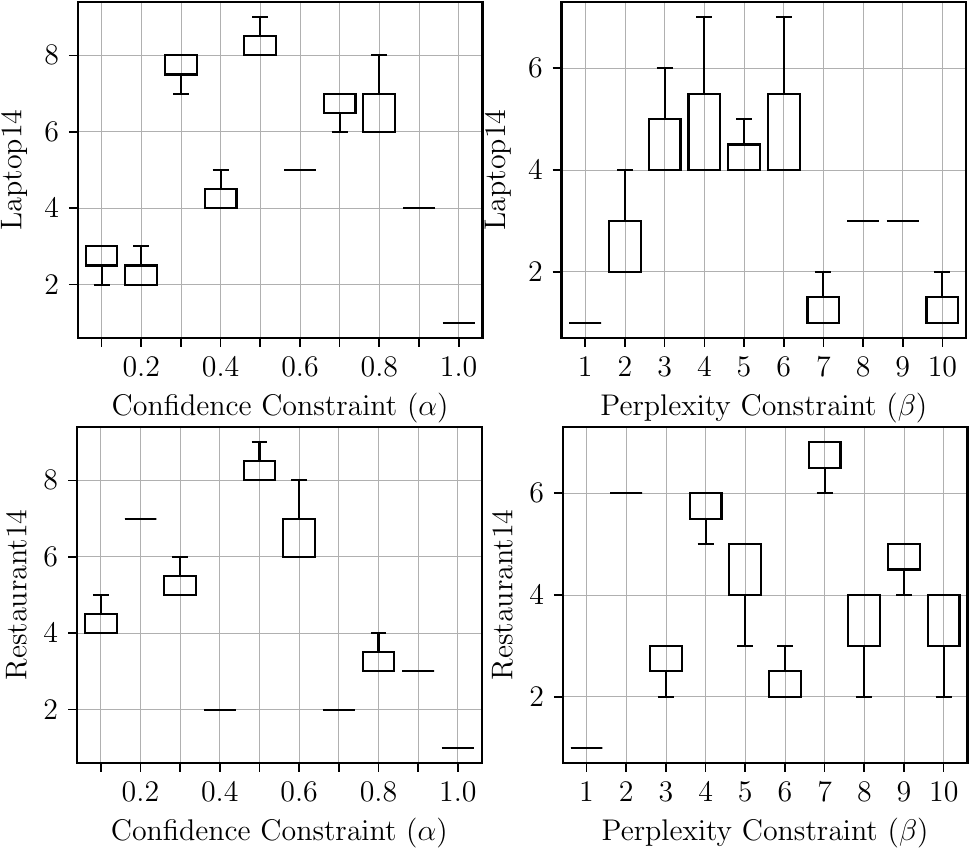}
    \caption{The Scott-knott rank test plots under different $\alpha$ and $\beta$ in \our (\texttt{EDA}). The bigger rank means better performance.}
    \label{fig:sk_rank}
\end{figure}

We find that there is no single best setting for the two hyper-parameters, $\alpha$ and $\beta$, in different situations such as different datasets and backend augmentation methods. To explore the sensitivity of these hyper-parameters, we conducted experiments on the \texttt{Laptop14} and \texttt{Restaurant14} datasets and show the Scott-Knott rank test ~\citep{MittasA13} plots and performance box plots in \pref{fig:sk_rank} and \pref{fig:box_plot}, respectively. We found that the best value of $\alpha$ highly depends on the dataset. For the \texttt{Laptop14} and \texttt{Restaurant14} datasets, a value of $\alpha=0.5$ was found to be the best choice according to \pref{fig:sk_rank}. However, it's worth noting that the smaller the value of $\alpha$, the higher the computation complexity due to the need for more augmentation instances. To balance efficiency and performance, we recommend a value of $\alpha=0.99$ ($\alpha=1$ means no augmentation instances survive) in \our, which reduces computation complexity. Additionally, we found that $\beta$ is relatively easy to determine, with a value of $\beta=4$ being commonly used.

%!TeX root=main.tex
\section{Related Works}
\label{sec:related_works}
As pretraining has advanced, text augmentation techniques have become an increasingly popular area of research~\citep{SennrichHB16,Coulombe18,LiJDLW19,WeiZ19,KumarCC20,LewisLGGMLSZ20,XieDHL020,BiLH20,RenZLSZ21,HaralabopoulosT21,WangSHTGJ22,YueLCYSY22,ZhouZTJY22,KamallooR22,WangZTWFJ0D22}. Many of these techniques focus on low-resource scenarios~\citep{ChenYY20,ZhouZTJY22,KimWOCH22,ZhouLHBCSM22,WuGLZH22,YangDCSRD22,WangXSHTGJ22,YangDCSRD22}. However, they tend to fail when applied to large public datasets \citep{ZhouZTJY22}. Recent prominent works~\cite{SennrichHB16,KumarCC20,LewisLGGMLSZ20,NgCG20,BodyTLLZ21,ChangSZDS21,LuoLLZ21,WangXSHTGJ22} recognize the significance of pre-trained language models (PLMs) for text augmentation and propose PLM-based methods to improve text augmentation. However, the quality of augmentation instances generated by unsupervised PLMs cannot be guaranteed. Some research \cite{ZFXM20} has attempted to use adversarial training in text augmentation, which can improve robustness, but these methods are more suitable for low-sample augmentation scenarios and cause shifted feature spaces in large datasets.

While recent studies have emphasized the importance of quality control for augmentation instances \citep{LewisWLMKPSR21,KamallooR22,WangXSHTGJ22}, there remains a need for a transferable augmentation instance-filtering framework that can serve as an external quality controller to improve existing text augmentation methods.

Our work aims to address the failure mode of large dataset augmentation and improve existing augmentation methods more widely. Specifically, \our\ is a simple but effective framework that can work with a variety of existing augmentation backends, including \texttt{EDA} \cite{WeiZ19} and PLM-based augmentation \cite{KumarCC20}.

\section{Conclusion}
Existing text augmentation methods usually lead to performance degeneration in large datasets due to numerous low-quality augmentation instances, while the reason for performance degeneration has not been well explained. We find low-quality augmentation instances usually have shifted feature space compare to natural instances. Therefore, we propose a universal augmentation instance filter framework \our\ to widely enhance existing text augmentation methods. \our\ is an external and flexible framework, all the existing text augmentation methods can be seamless improved. Experimental results on three TC datasets and five ABSC datasets show that \our\ is able to alleviate feature space shift in augmentation instances and significantly improve existing augmentation methods. 

\section*{Acknowledgements}
This work was supported by UKRI Future Leaders Fellowship (MR/X011135/1, MR/S017062/1), NSFC (62076056), Alan Turing Fellowship, EPSRC (2404317), Royal Society (IES/R2/212077) and Amazon Research Award.

\section{Limitations}
We propose and solve the feature space shift problem in text augmentation. However, there is a limitation that remains. \our\ cannot preserve the grammar and syntax to a certain extent. We apply the perplexity filtering strategy, but it is an implicit constraint and cannot ensure the syntax quality of the augmentation instances due to some breaking transformations, such as keyword deletions and modifications. However, we do not need precise grammar and syntax information in most classification tasks, especially in PLM-based classification. For some syntax-sensitive tasks, e.g., syntax parsing and the syntax-based ABSC~\cite{ZhangLS19,PhanO20,DaiYSLQ21}, ensuring the syntax quality of the augmented instances is an urgent problem. Therefore, \our\ may not be an best choice for some tasks or models requiring syntax as an essential modeling objective~\cite{ZhangLS19}. In other words, the syntax quality of \our\ depends on the backend.

\bibliography{ref}
\bibliographystyle{acl_natbib}

\appendix

\newpage
\section{Hyperparameter Settings}
\label{app:parameters}
\subsection{Hyperparameter Settings for \our}
Some important parameters are set as follows.
\begin{itemize}[leftmargin=*,noitemsep,nolistsep]
    \item $k$ is set to $5$ for the $k$-fold cross-boosting on all datasets. 
    \item The number of augmentation instances per example $\tilde{N}$ is $8$. 
    \item The transformation probability of each token in a sentence is set to $0.1$ for all augmentation methods.
    \item The fixed confidence and perplexity thresholds are set as $\alpha=0.99$ and $\beta=5$ based on grid search. We provide sensitivity test of $\alpha$ and $\beta$ in \pref{app:param}.
    \item The learning rates of base models \texttt{LSTM} and \texttt{DeBERTa-BASE} are set as $10^{-3}$ and $10^{-5}$, respectively. 
    \item The batch size and maximum sequence modeling length are $16$ and $80$, respectively.
    \item The $L_2$ regularization parameter $\lambda$ is $10^{-8}$; we use Adam as the optimizer for all models during the training process.
\end{itemize}

\section{Baseline Backends}
\label{app:backends_app}

We use \our\ to improve five state-of-the-art baseline text augmentation methods, all of which are used as the text augmentation back end of \our. Please refer to \pref{tab:rq4_results} for detailed experimental results.
\begin{itemize}[leftmargin=*,noitemsep,nolistsep]
    \item \texttt{EDA}(TextAttack\footnote{\url{https://github.com/QData/TextAttack}})~\cite{WeiZ19} performs text augmentation via random word insertions, substitutions and deletions.
    \item \texttt{SynonymAug}(NLPAug\footnote{\url{https://github.com/makcedward/nlpaug}})~\cite{NiuB18} replaces words in the original text with their synonyms. This method has been shown to be effective in improving the robustness of models on certain tasks.
    \item \texttt{SpellingAug}~\cite{Coulombe18}: it substitutes words according to spelling mistake dictionary.
    \item \texttt{SplitAug}~\cite{LiJDLW19} (NLPAug): it splits some words in the sentence into two words randomly.
    \item \texttt{BackTranslationAug}~\cite{SennrichHB16} (NLPAug): it is a sentence level augmentation method based on sequence translation.
    \item \texttt{ContextualWordEmbsAug}~\cite{KumarCC20} (NLPAug): it substitutes similar words according to the PLM (i.e., \texttt{Roberta-base}~\cite{LiuOGDJCLLZS19}) given the context.
\end{itemize}

\section{Additional Experiments}

\subsection{Natural Language Inference Experiments}
The experimental results in \pref{tab:nli} show that the performance of both BERT and DeBERTa models can be improved by applying \our. With \our, the accuracy of the BERT model on SNLI improves from 70.72\% to 73.08\%, and on MNLI from 51.11\% to 52.49\%. The DeBERTa model also shows significant improvement with EDA, achieving 86.39\% accuracy on SNLI and 78.04\% on MNLI. These results demonstrate the effectiveness of \our\ in improving the generalizability of natural language inference models, and its compatibility with different state-of-the-art pre-trained models such as BERT and DeBERTa.

\begin{table}[htbp]
  \centering
  \caption{The additional experimental results on the SNLI and MNLI datasets for natural language inference. The back end of \our\ is EDA.}
  \resizebox{\linewidth}{!}{
    \begin{tabular}{|c|c|c|c|c|c|}
    \hline
    \multirow{2}[1]{*}{\textbf{Augmentation}} & \multirow{2}[1]{*}{\textbf{Model}} & \multicolumn{2}{c|}{\textbf{SNLI}} & \multicolumn{2}{c|}{\textbf{MNLI}} \\
\cline{3-6}          &       & \texttt{Acc}   & \texttt{F1}    & \texttt{Acc}   & \texttt{F1} \\
    \hline
    \multirow{2}[1]{*}{\texttt{None}} & \texttt{BERT}  & 70.72 & 72.8  & 51.11 & 50.47 \\
\cline{2-6}          & \texttt{DeBERTa} & 83.50  & 83.47 & 74.75 & 74.62 \\
    \hline
    \multirow{2}[1]{*}{\our} & \texttt{BERT}  & 73.08 & 71.57 & 52.49 & 50.91 \\
\cline{2-6}          & \texttt{DeBERTa} & 86.39 & 86.16 & 78.04 & 77.04 \\
    \hline
%     \multirow{2}[1]{*}{\texttt{SynonymAug}} & \texttt{BERT}  &       &       &       &  \\
% \cline{2-6}          & \texttt{DeBERTa} &       &       &       &  \\
    \end{tabular}%
    }
  \label{tab:nli}%
\end{table}%

\begin{table*}[htbp]
    \centering
    \caption{Performance comparison of \our\ based on different augment back ends.}
    \scriptsize
    \resizebox{.75\linewidth}{!}{
      \begin{tabular}{c|c|c|c|c|c|c|c|c|c}
            \hline
            \multirow{1}[4]{*}{\textbf{Backend}} & \multirow{1}[4]{*}{\textbf{Model}} & \multicolumn{2}{c|}{\textbf{MAMS}} & \multicolumn{2}{c|}{\textbf{SST2}} & \multicolumn{2}{c|}{\textbf{SST5}} & \multicolumn{2}{c}{\textbf{AGNews10K}}\\
            \cline{3-10}          &       & \texttt{Acc}   & \texttt{F1}    & \texttt{Acc}   & \texttt{F1}    & \texttt{Acc}   & \texttt{F1} & \texttt{Acc}   & \texttt{F1} \\
            \hline
            \multirow{2}[3]{*}{\texttt{None}} 
            & \texttt{LSTM} & 56.96 & 56.18 & 82.37 & 82.37 & 44.39 & 43.60 & 87.60 & 87.36 \\
            & \texttt{BERT}  & 82.78 & 82.04 & 90.77 & 90.76 & 52.90 & 53.02  & 92.47 & 92.26\\
            & \texttt{DeBERTa}  & 83.31 & 82.87 & 95.28 & 95.28 & 56.47 &55.58  & 92.30 & 92.13 \\
            \hline
            \multirow{2}[3]{*}{\texttt{EDA}}
            & \texttt{LSTM}  &59.21 & 59.58 & 85.83 & 85.83 & 45.93 & 43.59 & 88.45 & 88.16 \\
            & \texttt{BERT}  & 84.01 & 83.44 & 92.33 & 92.33 & 53.94 & 52.80 & 92.48 & 92.25\\
            & \texttt{DeBERTa} & \textbf{84.51} & \textbf{83.97} & \textbf{96.09} & \textbf{96.09} & 57.78& \textbf{56.15} & 92.95 & 92.76 \\
            \hline
            \multirow{2}[3]{*}{\texttt{SpellingAug}} 
            & \texttt{LSTM}  & 58.50 & 57.65 & 85.23 & 85.23 & 43.39 & 42.45 & 87.93 & 87.63 \\
            & \texttt{BERT}  & 83.23 & 82.70 & 92.01 & 92.01 & 52.26 & 51.03 & 91.82 & 91.59  \\
            & \texttt{DeBERTa}  & 83.98 & 83.44 & 95.22 & 95.22 & \textbf{57.91} & 55.88  & 92.77 & 92.54 \\
            \hline
            \multirow{2}[3]{*}{\texttt{SplitAug}}
            & \texttt{LSTM}  & 58.65 & 57.23 & 85.64 & 85.64 & 46.04 & 43.97  & 87.65 & 87.42\\
            & \texttt{BERT}  & 83.05 & 82.49 & 92.20 & 92.20 & 51.86 & 51.39  & 91.92 & 91.69\\
            & \texttt{DeBERTa}  &82.67 & 82.26 & 94.76 &  94.76 & 57.67 & 55.90 & 92.70 & 92.51 \\
            %		\hline
            %		\multirow{2}[2]{*}{\texttt{BackTransAug}}
            %		& LSTM  &  &  &  &  &  &  \\
            %		 & BERT    &       &       & 91.35 & 91.35 &       &  \\
            %		& DeBERTa  & 83.27 & 82.77 & 95.22 & 95.22 & 56.31 & 55.1 \\
            \hline
            \multirow{2}[3]{*}{\texttt{WordEmdsAug}}
            & \texttt{LSTM}  & 59.54 & 57.58 & 86.30 & 86.30 & 46.47 & 44.15  & 88.38 & 88.10\\
            & \texttt{BERT}  & 83.31 & 82.72 & 91.76 & 91.76 & 52.49 & 50.27  & 92.43 & 92.24\\
            & \texttt{DeBERTa} & 83.35 & 82.87 & 95.33 & 95.33 & 57.22 & 56.08 & \textbf{93.88} & \textbf{93.70}\\
            \hline
        \end{tabular}
    }
    \label{tab:rq4_results}
\end{table*}

\begin{figure}[htbp]
    \centering
    \includegraphics[width=\linewidth]{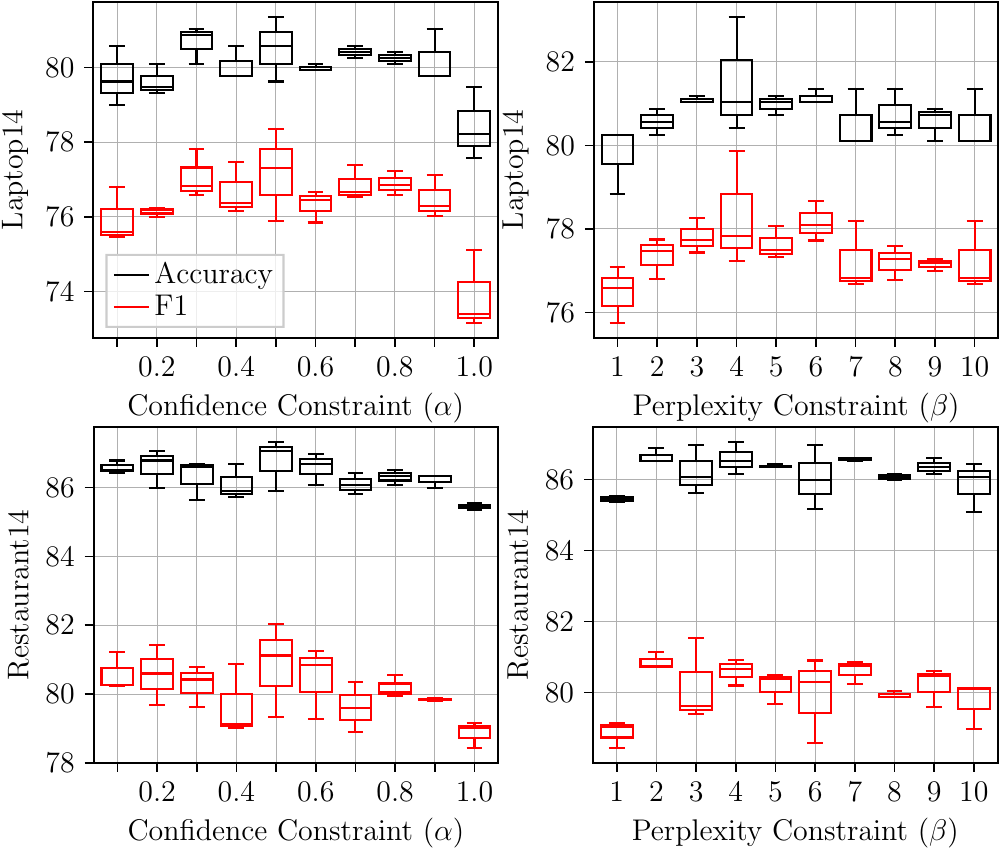}
    \caption{The performance box plots under different $\alpha$ and $\beta$ in \our (\texttt{EDA}).}
    \label{fig:box_plot}
\end{figure}

\subsection{Hyper-parameter Sensitivity Experiment}
\label{app:param}
We provide the experimental results of \our\ on the \texttt{Laptop14} and \texttt{Restaurant14} datasets in \pref{fig:box_plot}.

\subsection{Performance of \our\ on Different Backends}
\label{sec:rq4}

To investigate the generalization ability of \our, we evaluate its performance based on the existing augmentation backends. From the results shown in~\pref{tab:rq4_results}, we find that the performance of these text augmentation back ends can be improved by using our proposed \our. Especially by cross-referencing the results shown in~\pref{tab:full_experiments}, we find that the conventional text augmentation methods can be enhanced if appropriate instance filtering strategies are applied.

Another interesting observation is that PLMs are not effective for text augmentation, e.g., \texttt{WordEmdsAug} is outperformed by \texttt{EDA} in most comparisons\footnote{In fact, we also tried some other PLM-based augmentation back ends, e.g., \texttt{BackTranslationAug}, and we come up with same observation.}. Moreover, PLMs are resource-intense and usually cause a biased feature space. This is because PLMs can generate some unknown words, which are outside the testing set, during the pre-training stage. Our experiments indicate that using PLM as an augmentation instance filter, instead of a text augmentation tool directly, can help alleviate the feature space shift.

\subsection{Visualization of feature space}
\label{sec:app-tsne}
\begin{figure*}[t!]
\centering
\includegraphics[width=0.85\linewidth]{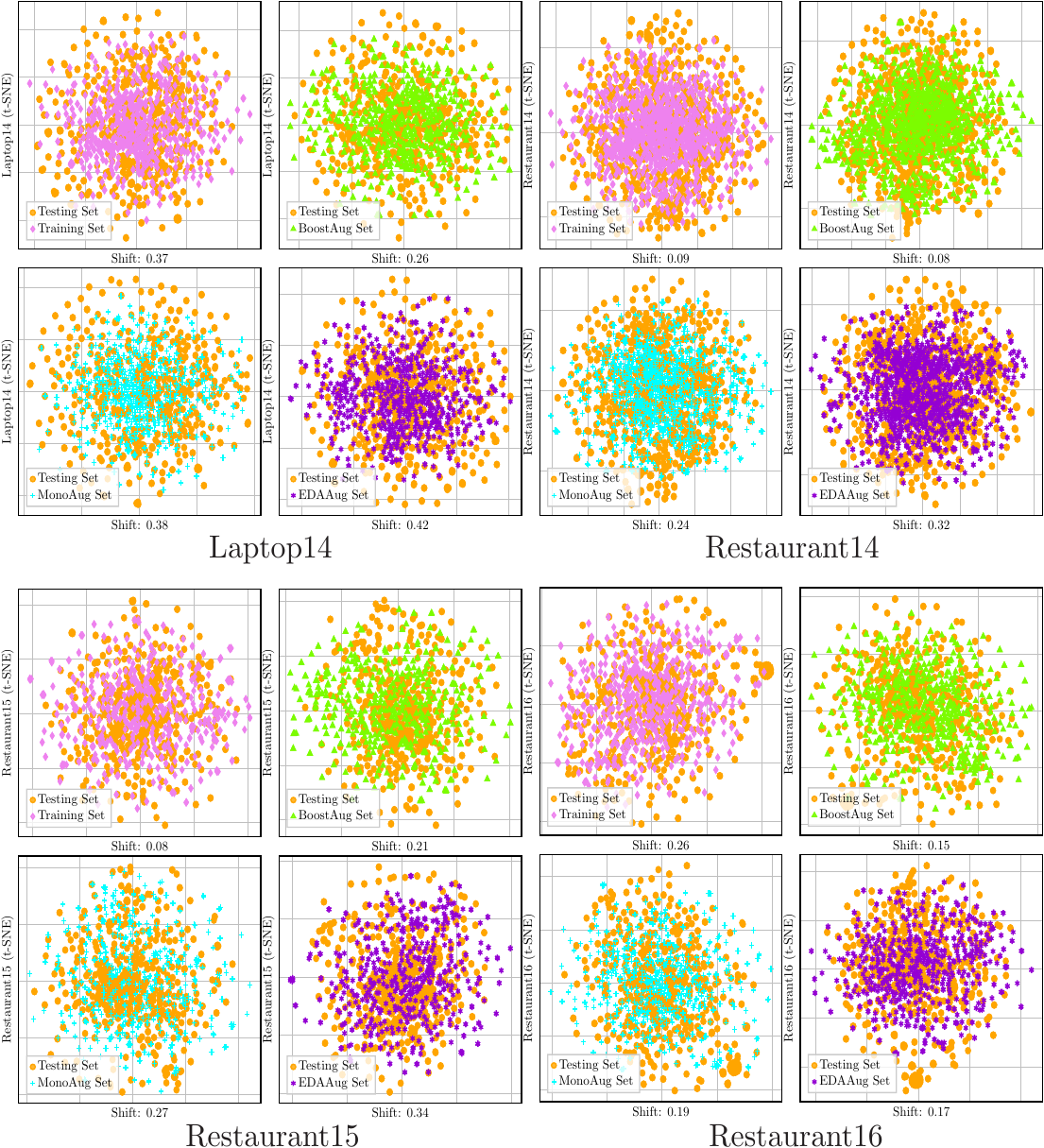}
\caption{
This figure shows the feature space shift ($\mathcal{S}$) of four ABSC datasets as visualized by $t$-SNE. The results demonstrate that \our\ has the least feature space shifts in comparison to other augmentation methods, such as \texttt{MonoAug} and \texttt{EDA}.}
\label{fig:app_tsne}
\end{figure*}

\pref{fig:app_tsne} shows the feature space shift of the ABSC datasets, where the augmentation back end of \our\ is \texttt{EDA}.

\subsection{Trajectory Visualization of RQ4}
\begin{figure*}[t!]
    \centering
    \includegraphics[width=1.\linewidth]{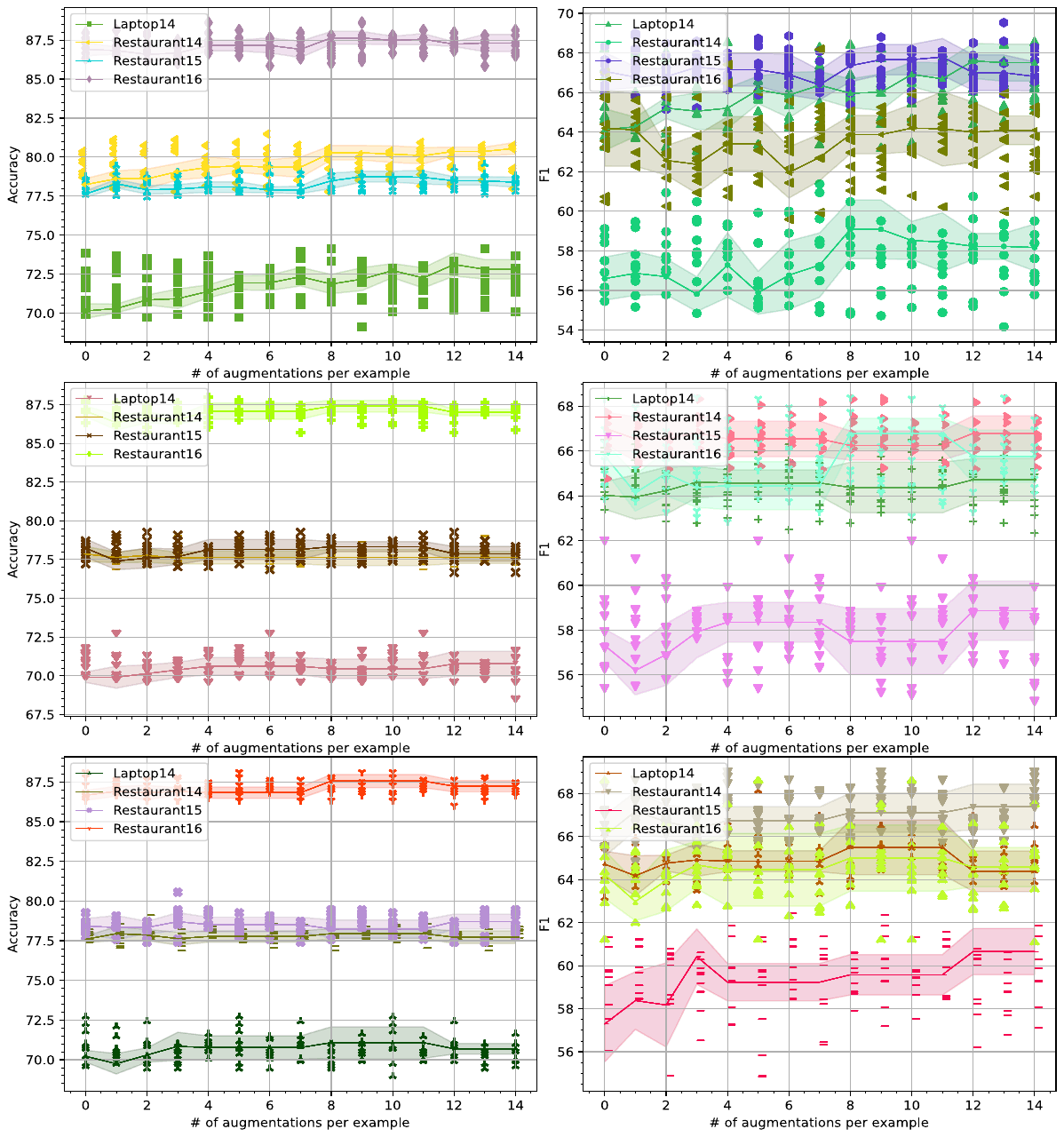}
    \caption{The performance (i.e., classification accuracy and F1 score) visualization of how \our\ perform as the number of augmentation instances per example increases. }
    \label{fig:rq4_full}
\end{figure*}

\pref{fig:rq4_full} shows the performance trajectory visualization of \texttt{MonoAug} and \texttt{EDA}. Compared to \our, \texttt{MonoAug} and existing augmentation methods usually trigger performance sacrifice while augmentation instances for each example are more than $3$.

\end{document}